\documentclass[11pt,a4paper]{article}
\usepackage{fullpage}

\usepackage{times}
\usepackage{latexsym}
\usepackage{amsfonts}
\usepackage{multirow}
\usepackage{hyperref}
\hypersetup{
  colorlinks = true,
  citecolor = blue
}
\usepackage{url}
\usepackage{booktabs}
\usepackage{mathtools}
\usepackage{enumitem}
\usepackage{color}
\usepackage{amsmath}
\usepackage{amssymb}
\usepackage{natbib}
\usepackage{xcolor}
\usepackage{nicefrac}

\usepackage{graphicx} 
\usepackage{epsfig} 
\usepackage{subfigure} 
\usepackage{caption}

\usepackage{algorithm}
\usepackage{algorithmic}

\usepackage{microtype}

\usepackage{breakurl}

\newcommand{\thead}[1]{\multicolumn{1}{c}{#1}}

\def\R{\mathbb{R}}

\def\O{\mathbb{O}}

\newcommand{\norm}[1]{\left\|#1\right\|}

\def\bx{{\mathbf x}}

\def\bz{{\mathbf z}}

\def\bB{\mathbf B}

\def\bU{{\mathbf U}}

\def\bW{{\mathbf W}}
\def\bX{{\mathbf X}}
\def\bY{{\mathbf Y}}

\def\minop{\mathop{\rm min}\limits}

\begin{document} 

\title{A Simple Approach to Learning\\ Unsupervised Multilingual Embeddings}

\author{Pratik Jawanpuria \qquad Mayank Meghwanshi \qquad Bamdev Mishra\\ $\ $ \\ Microsoft India\\         {Emails: $\{$pratik.jawanpuria, mamegh, bamdevm$\}$@microsoft.com}}

\date{}

\maketitle

\begin{abstract} 
Recent progress on unsupervised learning of cross-lingual embeddings in bilingual setting has given impetus to learning a shared embedding space for several languages without any supervision. A popular framework to solve the latter problem is to jointly solve the following two sub-problems: 1) learning unsupervised word alignment between several pairs of languages, and 2) learning how to map the monolingual embeddings of every language to a shared multilingual space. 
In contrast, we propose a two-stage framework in which we decouple the above two sub-problems and solve them separately using  existing techniques. 
Though this seems like a simple baseline approach, we show that the proposed approach obtains surprisingly good performance in various tasks such as bilingual lexicon induction, cross-lingual word similarity, multilingual document classification, and multilingual dependency parsing. 
When distant languages are involved, the proposed solution illustrates robustness and outperforms existing unsupervised multilingual word embedding approaches. 
Overall, our experimental results encourage development of multi-stage models for such challenging problems. 
\end{abstract}

\section{Introduction}\label{sec:intro}
Learning cross-lingual word representations has been the focus of many recent works  \citep{klementiev12a,mikolov13a,faruqui14a,ammar16,artetxe16a,conneau18a}. 
It aims at learning a shared embedding space for words across two (bilingual word embeddings or BWE) or more languages (multilingual word embeddings or MWE), by mapping similar words (or concepts) across different languages close to each other in the common embedding space. Such a representation is useful in various applications such as cross-lingual text classification \citep{klementiev12a,ammar16}, building bilingual lexicons \citep{mikolov13a}, cross-lingual information retrieval \citep{vulic15a}, sentiment analysis \citep{zhou15}, and machine translation \citep{gu2018,lample18a,artetxe18c}, to name a few. 

\citet{mikolov13a} observe that word embeddings exhibit similar structure across different languages. In particular, they show that the geometric arrangement of word embeddings can be (approximately) preserved by linearly transforming the word embeddings from one language space to another.  
Thereafter, several works have explored learning BWEs in supervised \citep{xing15a,artetxe16a,artetxe18a,smith17b,mjaw19a}  as well as unsupervised \citep{zhang17a,zhang17b,conneau18a,artetxe18b,alvares18a,hoshen18,grave18} settings. 
Supervision is usually in the form of a bilingual lexicon. Unsupervised BWE approaches enjoy the advantage of not requiring a bilingual lexicon during the training stage.  

Generalization from bilingual  to multilingual setting is desirable in various multilingual applications such as document classification, dependency parsing \citep{ammar16}, machine translation, etc. Representing the word embeddings of several languages in a common shared space can allow knowledge transfer across languages wherein a single classifier may be learned on multilingual datasets \citep{heyman19a}. 

In this work, we propose a two-stage framework for learning a shared multilingual space in the unsupervised setting. The two stages aim at solving the following  sub-problems: a) learning unsupervised word alignment between a few pairs of languages, and \textit{subsequently} b) learning how to map the monolingual embeddings of each language to a shared multilingual space. 
The sub-problems are separately solved using existing techniques. 
In contrast, existing state-of-the-art unsupervised MWE approaches \citep{chen18a,heyman19a,alaux19a} aim at solving the above sub-problems jointly. 
Though it appears like a simple baseline approach, the proposed framework provides the robustness and versatility often desirable while learning an effective shared MWE space for distant languages. Unsupervised alignment often fail for distant languages \citep{sogaard18,glavas19a,vulic19a}. 
Our approach can seamlessly work in ``hybrid'' settings, where supervision is available for some languages (but not for others). 
Integrating such hybrid setups in the joint optimization framework of existing unsupervised MWE approaches \citep{chen18a,heyman19a,alaux19a} may not\footnote{\citet{heyman19a}, for example, state that their approach is impractical in the supervised setting as it requires pairwise dictionaries for all pair of languages.} always be feasible. 

We evaluate our approach on different tasks such as bilingual lexicon induction (BLI), cross-lingual word similarity, and two downstream multilingual tasks: document classification and dependency parsing. We summarize our findings below.
\begin{itemize}
\item For a group consisting of similar languages, we observe that the proposed approach achieves BLI score similar to the existing unsupervised MWE approaches. 
We also observe that all multilingual approaches, including ours, benefit from transfer learning  across languages in such a setting.  
\item In a group comprising of distant languages, we observe sub-optimal BLI performance of existing unsupervised MWE approaches. The presence of distant languages sometimes adversely affect the alignment of even similar language pairs in such methods. The proposed approach, however, is robust and outperforms other multilingual approaches in such settings. 
\item The proposed approach performs better than other MWE baselines on cross-lingual word similarity, document classification, and dependency parsing tasks. 
\end{itemize}

The rest of the paper is organized as follows. 
Section~\ref{sec:proposed} discusses the proposed unsupervised MWE framework. Section~\ref{sec:experimentalsetup} describes our experimental setup and Section~\ref{sec:results} presents and analyzes the empirical results. Section~\ref{sec:conclusion} concludes the paper. We begin by discussing related works in the next section. 







\section{Related Work}\label{sec:relatedworks}
\textbf{Bilingual setting:} The problem of learning bilingual mapping of word embeddings in the supervised setup is as follows. 
Let $\bx_i\in\R^d$ and $\bz_j\in\R^d$ denote the $i$-th and $j$-th word embeddings of the source and the target languages, respectively. 
A seed translation dictionary $\bY\in\{0,1\}^{n\times n}$ is available during the training phase such that $\bY(i,j)\coloneqq y_{ij}=1$ if $\bx_i$ corresponds to $\bz_j$ and $y_{ij}=0$ otherwise. The aim is to learn a mapping from the source language embedding space to the target language embedding space. 
A popular approach to learn this mapping $\bW\in\R^{d\times d}$ is by solving the following Orthogonal Procrustes problem:
\begin{equation}\label{eqn:procrustes}
\minop_{\bW\in\O^d} \sum_{i,j} y_{ij}\norm{\bW\bx_i-\bz_j}^2,
\end{equation}
where $\O^d$ is the set of $d$-dimensional orthogonal matrices. 
Problem (1) admits a closed-form solution \citep{schonemann1966a}. Several improvements of above have been explored in this direction including pre/post-processing of embeddings \citep{artetxe18a}, additional transformations \citep{doval18a}, different loss functions \citep{joulin18b,mjaw19a}, to name a few. 

In the unsupervised setting, the seed dictionary (matrix $\bY$) is unknown. Popular unsupervised BWE frameworks include self-learning \citep{artetxe18b}, adversarial \citep{zhang17a,conneau18a}, and optimal transport \citep{zhang17b,alvares18a} based approaches. 
Most unsupervised algorithms aim at learning both $\bY$ and $\bW$ simultaneously, in a joint optimization framework. 
A few such the Gromov-Wasserstein approach \citep{peyre16a,alvares18a} explicitly aim at learning only the bilingual word alignment, i.e., the $\bY$ matrix,  and they suggest using (\ref{eqn:procrustes}) to learn the mapping operator ($\bW$). A few recent works have analyzed the effectiveness of supervised and unsupervised BWE approaches \citep{sogaard18,glavas19a,vulic19a}.

\noindent\textbf{Multilingual setting:} In the supervised setting, a popular multilingual approach is to bilingually map the embeddings of all other languages to a chosen pivot language \citep{ammar16,smith17b}. \citet{kementchedjhieva18a} propose to employ the generalized Procrustes analysis method \citep{gower75} to learn a MWE space. This, however, requires $n$-way dictionary for $n$ languages, which is a stringent constraint in real-world applications. 
Recently, \citet{mjaw19a} have proposed a geometric approach for learning a latent shared MWE space using only a few bilingual dictionaries ($n-1$ bilingual dictionaries for $n$ languages suffice). 

\citet{chen18a} are among the first to propose learning MWE in the unsupervised setting. They extend the unsupervised BWE approach of \citep{conneau18a} -- adversarial training and iterative refinement procedure -- to the multilingual setting. 
However, the GAN-based procedure of \citet{conneau18a} has known concerns related to optimization stability with distant language pairs \citep{sogaard18}. 
\citet{alaux19a} propose a joint optimization framework for learning word alignment ($\bY$) and mapping ($\bW$) between several pair of languages. It aims to learn the shared MWE space by optimizing direct mappings between pairs of languages as well as indirect mappings (via a pivot language). 
The bilingual alignments are learned as doubly-stochastic matrices and are modeled using the Gromov-Wasserstein loss function. The mapping operators are modeled using the non-smooth RCSLS loss function \citep{joulin18b}. For efficient optimization, they employ alternate minimization in a stochastic setting \citep{grave18}. \citet{heyman19a} propose to learn the shared MWE space by incrementally adding languages to it, one in each iteration. Their approach is based on a reformulation of the unsupervised BWE approach of \citet{artetxe18b}.

\section{Unsupervised Multilingual Multi-stage Framework}\label{sec:proposed}
We develop a two-stage algorithm for unsupervised learning of multilingual word embeddings (MWEs). The key idea is as follows:
\begin{itemize}
\item learn unsupervised word alignment between a few pairs of languages, and then
\item use the above (learned) knowledge to learn the shared MWE space.
\end{itemize}
We propose to solve the above two stages sequentially, using existing techniques. This is in contrast to the existing unsupervised MWE works  \citep{alaux19a,chen18a,heyman19a} that aim at learning the unsupervised word alignments and cross-lingual word embedding mappings jointly. 
Though the proposed approach appears simple, we empirically observe that it has better generalization ability and robustness. 
We summarize the proposed approach, termed as \textbf{U}nsupervised \textbf{M}ultilingual \textbf{M}ulti-stage \textbf{L}earning (UMML), in Algorithm~\ref{alg:geommgraph} and discuss the details below. 

\begin{algorithm}[t]
    \centering
    \caption{Unsupervised Multilingual Multi-stage Learning (UMML)}\label{alg:geommgraph}
    {
   \begin{tabular}{p{1\textwidth}} 
    {\bfseries Input:} Monolingual embeddings $\bX_i$ for each language $L_i$ and 
    an undirected, connected graph $G(V,E)$ with $V=\{L_1,\ldots,L_n\}$\\\\
    /*{\bfseries Stage 1:} Generating unsupervised bilingual lexicons $\bY_{ij}$*/ \\
     {\bfseries for }{each unordered pair  $(L_i,L_j)\in E$ }{\bfseries do}\\
     \ \ $\bY_{ij}\leftarrow$ \texttt{UnsupWordAlign}$(\bX_i,\bX_j)$ \\
     {\bfseries end for}\\\\
    /*{\bfseries Stage 2:} Learning multilingual word embeddings in a shared latent space*/\\
    Run \texttt{GeoMM} on $G(V,E)$ with monolingual embeddings $\bX_i$ for all languages $L_i$ and bilingual lexicons $\bY_{ij}$ for all language pairs $(L_i,L_j)\in E$ \\\\
    Output of \texttt{GeoMM}:\\
    a) metric $\bB$ (a positive definite matrix), and \\
    b) orthogonal matrices $\bU_i\ \forall i=1,\ldots,n$\\\\
    /*Representing word embedding $x$ of language $L_i$ in the common multilingual space*/\\
    $x\rightarrow \bB^\frac{1}{2}\bU_i^\top x $\\
   \end{tabular}
   }
\end{algorithm}


\subsection{Stage 1: Generating Bilingual Lexicons}
The first stage of our framework is to generate bilingual lexicons for a few pairs of languages. These lexicons are used in learning a shared MWE space in the second stage.  
We employ existing unsupervised bilingual word alignment algorithms \citep{artetxe18b,alvares18a} to generate the bilingual lexicons. 
It should be noted that these lexicons are learned in the bilingual setting independent of each other. Additionally, they can be learned in parallel. 
Our framework allows usage of different unsupervised bilingual word alignment algorithms \citep{artetxe18b,alvares18a,conneau18a} for different pairs of languages. 
More generally, bilingual lexicon for different pairs of languages may even be obtained using different class of algorithms/resources: unsupervised, weakly-supervised with bootstrapping \citep{artetxe17a}, human supervision, etc. 
This is because the second stage of our framework is agnostic of how the lexicons are obtained. Such flexibility in obtaining bilingual lexicons is desirable for learning a good quality shared MWE space for real-world applications since it has been observed that existing unsupervised bilingual word embedding algorithms may fail when languages are from distant families \citep{sogaard18,mjaw19a,glavas19a,vulic19a}. 
To the best of our knowledge, existing unsupervised MWE approaches do not discuss extensibility to such hybrid settings. 


We use two unsupervised bilingual word alignment algorithms \citep{alvares18a,artetxe18b} for generating bilingual lexicons, described in Section~\ref{subsec:implementation}. It should be emphasized that these bilingual lexicons are learned only for a few pairs of languages. Such pairs may be randomly chosen but should satisfy a very simple graph-connectivity criterion mentioned in the following section. In our experiments, $n-1$ bilingual lexicons are generated for $n$ languages. 


\subsection{Stage 2: Learning Multilingual Word Embeddings}\label{subsec:geomm}
As stated earlier, we learn the MWEs using the bilingual lexicons obtained from the first stage. To achieve our objective, we propose to employ the Geometry-aware Multilingual Mapping (GeoMM) algorithm \citep{mjaw19a}. 

The setting of GeoMM may be formalized as an undirected, connected graph $G$, whose nodes represent languages and edges between nodes imply availability of bilingual dictionaries (for the corresponding pair of languages). GeoMM represents multiple languages in a common latent space by learning language-specific rotations for each language ($d\times d$ orthogonal matrix $\bU_i$ for each language $L_i$) and a similarity metric common across languages (a $d\times  d$ symmetric positive-definite matrix $\bB$), where $d$ is the dimensionality of the monolingual word embeddings. 
The rotation matrices align the language embeddings to a common latent space, while the (common) metric $\bB$ governs how distances are measured in this latent space. 
Both the language-specific parameters ($\bU_i\ \forall L_i$) and the shared parameter ($\bB$) are learned via a joint optimization problem \citep[refer][Equation $3$]{mjaw19a}. 
The function that maps a word embedding $x$ from language $L_i$'s space to the common latent space is given by: $x\rightarrow \bB^\frac{1}{2}\bU_i^\top x$. 



\subsection{Implementation Details}\label{subsec:implementation}
We develop two variants of the proposed approach which differ only in the unsupervised bilingual word alignment algorithm employed in the first stage. 

\textbf{UMML-SL:} In this method, the \texttt{UnsupWordAlign} subroutine in Algorithm~\ref{alg:geommgraph} employs the unsupervised self-learning  algorithm developed by \citet{artetxe18b}. 
It should be noted that for UMML-SL, we do not employ various pre-processing and post-processing steps (such as whitening, de-whitening, symmetric re-weighting, etc.) that are included in the pipeline proposed by \citet{artetxe18b}. 
Hence, the simplified version of the self-learning algorithm used in our first stage only involves unsupervised initialization followed by stochastic dictionary induction \citep{artetxe18b}. 
This is done to ensure that all the compared approaches (for experiments discussed in Sections~\ref{sec:experimentalsetup} \& \ref{sec:results}) have the same input monolingual embeddings for their main algorithm. 

\textbf{UMML-GW:} 
We employ the Gromov-Wasserstein (GW) algorithm \citep{alvares18a} in the first stage of this method. 
The GW approach formulates the bilingual word alignment problem within the optimal transport framework \citep{memoli11a,peyre16a,peyre19a}. 
It learns a doubly stochastic matrix.
We then run a refinement procedure to obtain a bilingual word alignment \citep{conneau18a}. 
The key idea in the refinement procedure is that given the (probabilistic) word alignment obtained from the GW algorithm, a Procrustes cross-lingual mapping operator $\bW$ is learned by solving (\ref{eqn:procrustes}). The mapping operator $\bW$, in turn, is used to induce a (refined) bilingual lexicon $\bY$, using cross-domain similarity local scaling (CSLS) similarity measure \citep{conneau18a}. The vocabularies used for the refinement procedure is the same as those provided to the GW algorithm (i.e., top $20\,000$ most frequent words). 

\section{Experimental Setup}\label{sec:experimentalsetup}
The experiments are aimed the following key questions on learning multilingual embeddings via unsupervised word alignment: 
\begin{enumerate}
\item How does the proposed approach, learning bilingual word alignments and shared multilingual space sequentially, fare against existing methods that learn them jointly? 
\item How robust are the existing multilingual methods when distant languages are involved? Does the presence of a distant language affect the learning between two close-by languages?
\end{enumerate}

We perform rigorous evaluations on a number of tasks to answer the above questions.

\noindent\textbf{Bilingual lexicon induction (BLI):} A popular task to evaluate the learned cross-lingual mappings \citep{artetxe18b,chen18a,alaux19a,heyman19a}. We perform evaluations on the MUSE \citep{conneau18a} and the VecMap \citep{dinu15a,artetxe17a,artetxe18a} test datasets. Both the datasets contain pre-trained monolingual embeddings (but trained on different corpora). Both also provide test bilingual dictionaries for various language pairs involving English (en). MUSE also has dictionaries between a few non-English European languages: Spanish (es), French (fr), German (de), Italian (it), and Portuguese (pt). VecMap contains English to other language dictionaries for four languages: de, it, es, and Finnish (fi). MUSE contains more number of languages, many of which are included in our experiments: Arabic (ar), Bulgarian (bg), Czech (cs), Danish (da), Dutch (nl), Finnish(fi), Greek (el), Hindi (hi), Hungarian (hu), Polish (po), Russian (ru), and Swedish (sv). In a given dataset, if the test bilingual dictionary of a language pair is missing, (e.g., po-cs in MUSE), we follow \citep{alaux19a} and use the intersection of their full dictionaries with English (e.g., po-en and en-cs in MUSE) to construct a test set. Please refer \citep{dinu15a,artetxe18a} and \citep{conneau18a} for more details on VecMap and MUSE, respectively. 
Following existing works \citep{chen18a,heyman19a,alaux19a}, we report Precision$@1$ in the BLI experiments. 
For inference, we employ the cross-domain similarity local scaling (CSLS) score \citep{conneau18a} in the nearest neighbor search. 
The BLI results on the VecMap dataset are discussed in Appendix \ref{app:additional_results}. 

\noindent\textbf{Cross-lingual word similarity (CLWS):} We also evaluate the quality of multilingual word embeddings on the CLWS task using the SemEval 2017 dataset \citep{collados17}. 

One of the main goals for learning multilingual word embeddings is to enable transfer learning across languages for various downstream natural language applications. Hence, we also evaluate the methods on two other tasks:  multilingual document classification (MLDC) and multilingual dependency parsing (MLDP) \citep{ammar16,duong17a,heyman19a}. \citet{ammar16} provide a platform to evaluate MWEs on the two tasks. 

\noindent\textbf{MLDP:} We evaluate the quality of learned multilingual embeddings on MLDP dataset, MLParsing, sampled from the Universal Dependencies 1.1 corpus \citep{agic15a}. The dataset has twelve languages: Bulgarian, Czech, Danish, German, Greek, English, Spanish, Finnish, French, Hungarian, Italian, and Swedish. It has training and test sets, containing $6748$ and $1200$ sentences, respectively. While the test set for each language contains $100$ sentences, the sentences in the training set of each language vary from $98$ to $6694$. The stack-LSTM parser \citep{dyer15a} used in this setup is configured to not use any part-of-speech/morphology attributes and to keep the input word embeddings fixed \citep{ammar16}. 
\newline
\noindent\textbf{MLDC:} This task is evaluated on the ReutersMLDC document classification dataset, which has documents in seven languages: Danish, German, English, Spanish, French, Italian, and Swedish. The training and the test sets contains $7000$ and $13\,058$ documents, respective, well balanced across languages \citep{ammar16,heyman19a}. The document classifier is based on the average perceptron \citep{klementiev12a}. 


\textbf{Compared methods:} In addition to the proposed methods \textbf{UMML-GW} and \textbf{UMML-SL}, discussed in Section~\ref{subsec:implementation}, we consider other unsupervised multilingual word embeddings baselines: \textbf{UMWE} \citep{chen18a} and \textbf{UMH}  \citep{alaux19a}. 
For BLI experiments, we also evaluate state-of-the-art unsupervised bilingual word embeddings approach of \citet{artetxe18b}, \textbf{BilingUnsup}, to observe the effect of transfer learning in multilingual approaches. 
As in existing works, all the unsupervised methods use the top $20\,000$ most frequent words of each language to learn the shared embedding space. 
We use the same hyper-parameters for UMWE, UMH, and BilingUnsup as suggested by their authors. 

\begin{table*}[t]\centering
\setlength{\tabcolsep}{5pt}
{\footnotesize
\centering
\begin{tabular}{lrrrrrrrrrrrrr}
\toprule
 & \thead{de-xx} & \thead{en-xx}  & \thead{es-xx} & \thead{fr-xx}  & \thead{it-xx}  & \thead{pt-xx} & \thead{xx-de} & \thead{xx-en}  & \thead{xx-es} & \thead{xx-fr}  & \thead{xx-it} & \thead{xx-pt} & \thead{avg.} \\
\midrule 
BilingUnsup & $60.9$ & $76.9$ & $75.6$ & $72.7$ & $75.2$ & $75.3$ & $61.6$ & $76.1$ & $77.2$ & $75.9$ & $73.6$ & $72.2$ & $72.8$ \\
\midrule
UMML-GW    & $69.3$ & $80.2$ & $81.2$ & $78.9$ & $80.3$ & $79.9$ & $69.0$ & $\mathbf{81.7}$ & $81.7$ & $82.0$ & $78.7$ & $76.7$ & $78.3$\\
UMML-SL  & $\mathbf{70.5}$ & $80.0$ & $81.7$ & $79.7$ & $\mathbf{80.9}$ & $\mathbf{80.9}$ & $\mathbf{69.9}$ & $80.6$ & $82.3$ & $83.1$ & $79.6$ & $78.2$ & $\mathbf{79.0}$ \\
UMWE         & $70.4$ & $\mathbf{80.6}$ & $\mathbf{82.0}$ & $\mathbf{79.8}$ & $80.6$ & $80.6$ & $69.5$ & $77.4$ & $\mathbf{83.5}$ & $\mathbf{84.1}$ & $\mathbf{80.4}$ & $\mathbf{79.0}$ & $\mathbf{79.0}$\\ 
UMH            & $69.2$ & $79.9$ & $81.8$ & $79.4$ & $80.6$ & $80.6$ & $69.0$ & $80.7$ & $82.3$ & $82.8$ & $79.0$ & $77.6$ & $78.6$ \\ 
\bottomrule
\end{tabular}
}
\caption{
Average Precision$@1$ for BLI obtained by multilingual algorithms on six European languages (German, English, Spanish, French, Italian, and Portuguese) from the MUSE dataset. The results are obtained for every combination of source-target pair. The column de-xx (xx-de) denotes average performance when German is considered the source (target) language. It can be observed that all the approaches obtain similar generalization on the test set. The final average of each method is computed on thirty results.  
}\label{table:multilingual-six-euro}
\end{table*}

\begin{table*}\centering
\setlength{\tabcolsep}{5pt}
{\footnotesize
\centering
\begin{tabular}{lrrrrrrrrrrrr}
\toprule
 & \thead{cs-xx}  & \thead{da-xx} & \thead{de-xx}  & \thead{en-xx} & \thead{es-xx}  & \thead{fr-xx} & \thead{it-xx}  & \thead{nl-xx} & \thead{pl-xx}  & \thead{pt-xx} & \thead{ru-xx} \\
\midrule 
BilingUnsup & $61.8$ & $58.7$ & $58.4$ & $64.9$ & $65.0$ & $63.3$ & $64.5$ & $63.7$ & $62.0$ & $64.4$ & $59.3$ \\
\midrule
UMML-GW    & $64.6$ & $\mathbf{61.7}$ & $\mathbf{64.1}$ & $70.0$ & $\mathbf{69.3}$ & $68.0$ & $\mathbf{68.7}$ & $67.1$ & $65.6$ & $68.3$ & $62.4$ \\
UMML-SL      & $\mathbf{65.1}$ & $61.3$ & $\mathbf{64.1}$ & $\mathbf{70.2}$ & $\mathbf{69.3}$ & $\mathbf{68.1}$ & $\mathbf{68.7}$ & $\mathbf{67.4}$ & $\mathbf{66.0}$ & $\mathbf{68.7}$ & $\mathbf{63.3}$ \\
UMWE         & $57.6$ & $54.1$ & $56.8$ & $63.1$ & $62.9$ & $61.5$ & $61.9$ & $0.0$ & $58.6$ & $61.6$ & $56.3$ \\
UMH            & $63.7$ & $60.8$ & $62.8$ & $68.8$ & $68.9$ & $67.5$ & $68.0$ & $66.1$ & $64.2$ & $67.8$ & $61.9$ \\
\midrule
& \thead{xx-cs}  & \thead{xx-da} & \thead{xx-de}  & \thead{xx-en} & \thead{xx-es}  & \thead{xx-fr} & \thead{xx-it}  & \thead{xx-nl} & \thead{xx-pl}  & \thead{xx-pt} & \thead{xx-ru} & \thead{avg.} \\
\midrule 
BilingUnsup & $51.0$ & $56.6$ & $64.5$ & $69.6$ & $71.7$ & $70.1$ & $67.7$ & $66.1$ & $53.6$ & $68.8$ & $46.3$ & $62.4$ \\
\midrule
UMML-GW  & $53.1$ & $\mathbf{62.1}$ & $69.3$ & $74.6$ & $\mathbf{76.3}$ & $75.6$ & $72.3$ & $70.0$ & $55.1$ & $73.8$ & $47.6$ & $66.3$\\
UMML-SL  & $\mathbf{53.6}$ & $61.9$ & $\mathbf{69.5}$ & $\mathbf{75.0}$ & $\mathbf{76.3}$ & $\mathbf{75.7}$ & $\mathbf{72.4}$ & $\mathbf{70.1}$ & $\mathbf{55.2}$ & $\mathbf{74.0}$ & $\mathbf{48.4}$ & $\mathbf{66.6}$\\
UMWE         & $49.5$ & $57.6$ & $60.3$ & $63.1$ & $68.4$ & $67.9$ & $65.2$ & $0.0$ & $51.0$ & $66.3$ & $45.0$ & $54.0$\\
UMH            & $52.9$ & $60.4$ & $68.3$ & $74.1$ & $75.6$ & $74.6$ & $71.4$ & $68.6$ & $54.8$ & $72.6$ & $47.4$ & $65.5$\\
\bottomrule
\end{tabular}
}
\caption{Average Precision$@1$ for BLI on eleven European languages from the MUSE dataset. The languages are from different families such as Latin (Spanish, French, Italian, Portuguese), Germanic (Danish, German, English, Dutch), and Slavic (Czech, Polish, Russian). The results are obtained for every combination of source-target pair.  
The proposed methods, UMML-SL and UMML-GW, obtain better performance than UMH and UMWE in each column. This illustrates the robustness of the proposed  framework with respect to diversity in languages. 
The final average of each method is computed on $110$ results.}\label{table:multilingual-eleven-euro}
\end{table*}

\section{Results and discussion}\label{sec:results}
We now discuss and analyze the results obtained on the experimental setup described in Section \ref{sec:experimentalsetup}. 

\subsubsection*{Bilingual Lexicon Induction (BLI) Results}
\noindent\textbf{Experiment 1:} We begin with BLI task on a group of six relatively close European languages -- German, English, Spanish, French, Italian, and Portuguese -- from the MUSE dataset. We experiment on every pair of languages and in both directions, leading to thirty results for each method. Table~\ref{table:multilingual-six-euro} provides summarized results of this experiment. We observe that the proposed two-stage methods, UMML-GW and UMML-SL, obtain scores on par with state-of-the-art methods, UMWE and UMH. This shows that in case of similar/close-by languages, all the methods are able to learn a shared multilingual space with similar generalization performance. We also observe that all the multilingual methods outperform BilingUnsup, highlighting the benefit of transfer learning in this scenario. 

\noindent\textbf{Experiment 2:} We next expand the language set in the first experiment to include five other European languages (Czech, Danish, Dutch, Polish, Russian) from diverse language families (all from the MUSE dataset). This group of eleven languages has also been employed by \citet{alaux19a} in their BLI experiments. 
Table~\ref{table:multilingual-eleven-euro} reports the summarized results. We observe that the proposed two-stage methods, UMML-GW and UMML-SL, perform better than UMH and outperforms UMWE. In fact both the proposed methods obtain better results than UMH and UMWE in the every column of Table~\ref{table:multilingual-eleven-euro}. 
In this multilingual setting, UMWE fails to satisfactorily map the Dutch language word embeddings in the shared multilingual space, though Dutch is similar to English. It should be noted that in the bilingual setup UMWE learns an effective English-Dutch cross-lingual space (obtaining average en-nl and nl-en score of $75.2$). It, therefore, appears that UMWE has limitations in such multilingual settings which lead to its poor performance.


\begin{table*}[t]\centering
\setlength{\tabcolsep}{3.7pt}
{\footnotesize
\centering
\begin{tabular}{lrrrrrrrrrrrrrrrr}
\toprule
 & \thead{ar-de} & \thead{ar-en}  & \thead{ar-fr} & \thead{ar-hi}  & \thead{ar-ru}  & \thead{de-en} & \thead{de-fr} & \thead{de-hi}  & \thead{de-ru} & \thead{en-fr}  & \thead{en-hi} & \thead{en-ru} & \thead{fr-hi}  & \thead{fr-ru} & \thead{hi-ru} \\
\midrule 
BilingUnsup & $\mathbf{46.5}$ & $46.4$ & $55.0$ & $36.9$ & $\mathbf{35.2}$ & $70.8$ & $61.9$ & $31.8$ & $43.6$ & $79.8$ & $31.3$ & $44.1$ & $36.1$ & $44.9$ & $24.9$ \\
\midrule
UMML-GW & $\mathbf{46.5}$ & $\mathbf{50.5}$ & $58.1$ & $0.0$ & $33.6$ & $74.0$ & $\mathbf{75.5}$ & $0.0$ & $44.4$ & $\mathbf{82.5}$ & $0.0$ & $47.7$ & $0.0$ & $46.2$ & $0.0$ \\
UMML-SL   & $46.2$ & $49.5$ & $56.5$ & $39.4$ & $34.1$ & $74.6$ & $75.2$ & $\mathbf{38.4}$ & $\mathbf{45.2}$ & $\mathbf{82.5}$ & $\mathbf{39.0}$ & $\mathbf{49.7}$ & $\mathbf{42.7}$ & $\mathbf{47.1}$ & $\mathbf{29.7}$ \\
UMWE         & $0.0$ & $45.0$ & $\mathbf{58.4}$ & $\mathbf{41.4}$ & $0.0$ & $0.0$ & $0.0$ & $0.0$ & $40.2$ & $81.9$ & $36.4$ & $0.0$ & $42.6$ & $0.0$ & $0.0$\\ 
UMH            & $0.1$ & $0.1$ & $0.3$ & $0.0$ & $0.1$ & $\mathbf{74.7}$ & $72.5$ & $0.0$ & $44.7$ & $82.0$ & $0.0$ & $46.5$ & $0.0$ & $44.5$ & $0.0$\\ 
\midrule 
& \thead{de-ar} & \thead{en-ar}  & \thead{fr-ar} & \thead{hi-ar}  & \thead{ru-ar}  & \thead{en-de} & \thead{fr-de} & \thead{hi-de}  & \thead{ru-de} & \thead{fr-en}  & \thead{hi-en} & \thead{ru-en} & \thead{hi-fr}  & \thead{ru-fr} & \thead{ru-hi} & \thead{avg.}  \\
\midrule 
BilingUnsup & $30.8$ & $29.4$ & $37.7$ & $28.7$ & $\mathbf{35.0}$ & $72.0$ & $61.3$ & $42.0$ & $59.6$ & $78.7$ & $37.6$ & $59.2$ & $45.4$ & $62.6$ & $32.3$ & $46.7$ \\
\midrule
UMML-GW & $\mathbf{31.5}$ & $35.6$ & $37.5$ & $0.0$ & $32.8$ & $74.6$ & $ 70.5$ & $0.0$ & $61.3$ & $83.1$ & $0.0$ & $62.9$ & $0.0$ & $ 65.8$ & $0.0$ & $37.2$\\
UMML-SL   & $31.1$ & $35.5$ & $37.4$ & $\mathbf{29.7}$ & $33.9$ & $\mathbf{75.1}$ & $\mathbf{70.7}$ & $\mathbf{45.5}$ & $\mathbf{61.7}$ & $82.9$ & $\mathbf{47.6}$ & $\mathbf{65.6}$ & $\mathbf{51.9}$ & $\mathbf{66.6}$ & $\mathbf{39.9}$ & $\mathbf{50.8}$\\
UMWE         & $0.1$ & $\mathbf{37.6}$ & $\mathbf{39.8}$ & $23.7$ & $0.1$ & $0.0$ & $0.0$ & $0.0$ & $55.7$ & $ 79.5$ & $34.1$ & $0.0$ & $48.4$ & $0.0$ & $0.0$ & $22.2$\\ 
UMH            & $0.2$ & $0.1$ & $0.2$ & $0.0$ & $0.1$ & $ 74.3$ & $ 69.9$ & $ 0.0$ & $60.8$ & $\mathbf{83.2}$ & $0.0$ & $62.5$ & $0.0$ & $ 65.1$ & $0.0$ & $26.1$\\ 
\bottomrule
\end{tabular}
}
\caption{
Precision$@1$ for BLI obtained by multilingual algorithms on a highly diverse group of six languages (Arabic, German, English, French, Hindi, and Russian) from the MUSE dataset. 
The `avg' column reports the average performance of each method. 
Both the proposed methods, UMML-GW and UMML-SL, obtain better results than UMH and UMWE in this challenging BLI setup, with  UMML-SL outperforming every method. 
}\label{table:multilingual-diverse}
\end{table*}

\noindent\textbf{Experiment 3: } We next aim to learn the multilingual space for a more diverse group of languages from the MUSE dataset: Arabic, German, English, French, Hindi, and Russian. Table~\ref{table:multilingual-diverse} reports the BLI performance for each pair of languages. We observe that, except the proposed UMML-SL, all other multilingual methods fail to learn a reasonably good shared multilingual space for all languages. 
The proposed UMML-GW fails to obtain a reasonable BLI score ($<1$ Precision$@1$) in $10$ out of $30$ pairs. 
UMWE and UMH suffer from such failure on $16$ and $18$ pairs, respectively. 
On the other hand, the proposed UMML-SL obtains effective alignment on all pairs of languages, illustrating its robustness in such challenging setting. 
It obtains better performance than BilingUnsup, benefiting from transfer learning in such a diverse group as well. 
In the following, we analyze the results of the other three multilingual methods:\\
$\bullet$ The Gromov-Wasserstein (GW) alignment algorithm \citep{alvares18a}, which is used in the first stage of the proposed UMML-GW, does not learn an effective alignment of English and Hindi words. However, in its second stage, this ``misalignment" does not adversely affect the BLI performance of  language pairs not involving Hindi. This can be concluded from the observation that its BLI score for all language pairs not involving Hindi is similar to the corresponding scores obtained by UMML-SL. 
Overall, UMML-GW is able to learn suitable multilingual embeddings for five languages (ar, de, en, fr, and ru). \\
$\bullet$ UMH also employs GW based formulation \citep{alvares18a} in its joint optimization framework and it, too, does not learn suitable Hindi embeddings in the shared multilingual space. However, UMH also fails to learn suitable Arabic embeddings in the shared multilingual space. This is surprising since the GW algorithm (employed in the first stage of UMML-GW) learns an effective alignment of English and Arabic words. Hence, it appears that jointly learning the unsupervised alignment and multilingual mapping can adversely affect distant languages (Arabic in this case). Overall, UMH learns suitable multilingual embeddings for four languages (de, en, fr, and ru). \\
$\bullet$ The GAN-based approach, UMWE, learns two groups of alignment in the shared multilingual space. The first group consists of Arabic, English, French, and Hindi language embeddings, which are suitably aligned with each other. However, these are ``misaligned'' with German and Russian language embeddings in the shared space. 
On the other hand, the German and Russian language embeddings are suitably aligned with each other (but not with any other language). 
Such grouping cannot be attributed to language similarity (since English and German are closer than, for e.g., English and Arabic) and may result from optimizing instability \citep{sogaard18}. 


\begin{table*}[t]\centering
{\footnotesize
\centering
\begin{tabular}{lrrrrrrr}
\toprule
& \thead{en-de} & \thead{en-es}  & \thead{de-es} & \thead{en-it}  & \thead{de-it} & \thead{es-it} & \thead{avg.} \\
\midrule 
Luminoso  \citep{speer17}   & $0.769$ & $0.772$ & $0.735$ & $0.787$ & $0.747$ & $0.767$ & $0.763$ \\
NASARI \citep{collados16}       & $0.594$ & $0.630$ & $0.548$ & $0.647$ & $0.557$ & $0.592$ & $0.595$ \\
\midrule 
UMML-GW & $0.726$ & $0.724$ & $0.723$ & $0.728$ & $0.706$ & $0.742$ & $0.725$ \\
UMML-SL   & $0.725$ & $0.723$ & $0.721$ & $0.728$ & $0.705$ & $0.742$ & $0.724$ \\
UMWE         & $0.713$ & $0.706$ & $0.698$ & $0.708$ & $0.684$ & $0.727$ & $0.706$ \\
UMH            & $0.719$ & $0.721$ & $0.710$ & $0.727$ & $0.696$ & $0.732$ & $0.718$ \\
\bottomrule
\end{tabular}
}
\caption{Spearman correlation $(\rho)$ on the SemEval 2017 cross-lingual word similarity task. 
Among unsupervised methods, the proposed UMML-GW and UMML-SL perform better than UMWE and UMH. 
}\label{table:clws}
\end{table*}

\begin{table}\centering
{\footnotesize
\centering
\begin{tabular}{lrr}
\toprule
& \thead{MLDC} & \thead{MLDP}  \\
\midrule 
UMML-GW & $89.7$ & $69.9$\\
UMML-SL   & $\mathbf{90.3}$ & $\mathbf{71.0}$ \\
UMWE         & $88.3$ & $\mathbf{71.0}$ \\ 
UMH            & $90.0$ & $70.6$ \\
\bottomrule
\end{tabular}
}
\caption{
Average accuracy and unlabeled attachment score (UAS) across languages on ReutersMLDC (MLDC) and MLParsing (MLDP) datasets, respectively. The proposed UMML-SL obtain the best results in both downstream applications. 
}\label{table:downstream}
\end{table}

\subsubsection*{Cross-lingual Word Similarity (CLWS) Results}
\noindent\textbf{Experiment 4: } Table~\ref{table:clws} reports performance on the SemEval 2017 CLWS task for four languages: English, German, Spanish, and Italian. For evaluating the unsupervised MWE approaches, we consider the multilingual word embeddings of the four languages learned in BLI Experiment 2 (corresponding to Table~\ref{table:multilingual-eleven-euro}). 
Among the unsupervised MWE approaches, we observe that UMML-GW and UMML-SL obtain the best results. 
For reference, we also include the results of the SemEval 2017 baseline and the best reported system, NASARI \citep{collados16}  and Luminoso \citep{speer17}, respectively, in Table~\ref{table:clws}. However, it should be noted that both NASARI and Luminoso use additional knowledge sources like the Europarl and the OpenSubtitles2016 parallel corpora.

\subsubsection*{Results on Downstream Applications}
\noindent\textbf{Experiment 5: } For each of the four methods, we learn the shared multilingual space for twelve languages in the multilingual dependency parsing (MLDP) dataset: Bulgarian, Czech, Danish, German, Greek, English, Spanish, Finnish, French, Hungarian, Italian, and Swedish. 
We employ the $300$ dimensional pre-trained monolingual embeddings from the MUSE dataset, vocabulary list being the top $200\,000$ words in each language, as in the BLI experiments. 
The learned multilingual embeddings are also employed in the multilingual document classification (MLDC) evaluation, which has seven languages: Danish, German, English, Spanish, French, Italian, and Swedish. 
It should be noted that for this MLDP and MLDP tasks, related works \citep{ammar16,duong17a,heyman19a} trained $512$ dimensional monolingual embeddings on the datasets used by \citet{ammar16} and \citet{duong17a}. Hence, the presented results are not comparable with previously reported results. 

Table~\ref{table:downstream} reports the performance on the MLDP and MLDC tasks. We observe that the proposed two-stage approaches perform well on the downstream tasks with UMML-SL obtaining the best results.

\section{Discussion and Conclusion}\label{sec:conclusion}
In this work, we propose a two-stage framework for learning unsupervised multilingual word embeddings (MWE). The two stages correspond to first learning unsupervised word alignment between a few pairs of languages and subsequently learning a latent shared multilingual space. The two problem are solved using existing techniques: the first stage is solved using the self-learning \citep{artetxe18b} or the Gromov-Wasserstein alignment \citep{alvares18a} algorithms and the second stage is solved using the GeoMM algorithm \citep{mjaw19a}. 

Though the two-stage framework seems a simple approach compared to the existing joint optimization methods \citep{chen18a,alaux19a}, our main contribution has been to show that it is a strong performer. We observe that the proposed approach (UMML-SL) outperforms existing approaches in various tasks such as bilingual lexicon induction, cross-lingual word similarity, multilingual document classification, and multilingual document parsing. The proposed approach also exhibit robustness while learning the MWE space for highly diverse groups of languages, a challenging setting for existing approaches. 
The experiments were focused on the unsupervised setting, but the proposed two-stage framework has the necessary flexibility to be easily employed in fully supervised setting or in hybrid setups where supervision is available for some languages but may be unavailable for others. Our results encourage development of multi-stage models for learning multilingual word embeddings.


\bibliography{main}
\bibliographystyle{plainnat}

\clearpage

\appendix

\section{Additional results}\label{app:additional_results}

\begin{table*}[t]\centering
	\setlength{\tabcolsep}{5pt}
	{\footnotesize
		\centering
		\begin{tabular}{lrrrrrrrrrrr}
			\toprule
			& \thead{de-xx} & \thead{en-xx}  & \thead{es-xx} & \thead{fi-xx}  & \thead{it-xx} & \thead{xx-de} & \thead{xx-en}  & \thead{xx-es} & \thead{xx-fi}  & \thead{xx-it} & \thead{avg.} \\
			\midrule 
			UMML-SL   & $\mathbf{49.5}$ & $\mathbf{40.4}$ & $\mathbf{45.6}$ & $\mathbf{45.8}$ & $\mathbf{51.3}$ & $\mathbf{52.4}$ & $\mathbf{37.1}$ & $\mathbf{52.9}$ & $\mathbf{37.2}$ & $\mathbf{53.0}$ & $\mathbf{46.5}$\\
			UMWE         & $37.8$ & $0.2$ & $38.0$ & $35.1$ & $40.5$ & $39.4$ & $0.1$ & $42.8$ & $28.6$ & $40.7$ & $30.3$\\ 
			\bottomrule
		\end{tabular}
	}
	\caption{
		Average Precision$@1$ for BLI on five European languages (German, English, Spanish, Finnish, Italian) from the VecMap dataset. The results are reported for every combination of source-target pair. 
		The proposed method, UMML-SL, significantly improves the results of UMWE. 
	}\label{table:multilingual-five-euro-vecmap}
\end{table*}


\subsubsection*{Bilingual Lexicon Induction (BLI) Results}
\noindent\textbf{Experiment 6: } In addition to the MUSE dataset, we also perform BLI experiment on the VecMap dataset. Table~\ref{table:multilingual-five-euro-vecmap} (page 2 of this draft) reports result on VecMap for the proposed UMML-SL and UMWE methods. We observe that UMWE learns suitable alignments for only four languages (de, es, fi, and it). It is not able to effectively align the English language embeddings in the shared multilingual space. On the other hand, the proposed UMML-SL suitably aligns all the five languages, obtaining best performance in every summarized result (Table~\ref{table:multilingual-five-euro-vecmap}). 
On this dataset, we do not observe suitable alignment between pairs of language using the GW algorithm \citep{alvares18a}. This might be due to the challenging nature of this dataset, which \citet{alvares18a} mention. In their work, \citet{alvares18a} have alluded to some specific normalization performed on the covariance matrices of this dataset. The authors have not responded to our query on this. Hence, without the required normalization, both UMML-GW and UMH fail to learn a good alignment on any pair of language. \\

\noindent\textbf{Detailed results of Experiment 1:} The detailed results of Experiment 1 (of the main paper) are provided in Table~\ref{table:multilingual-european-detailed} in Page 2 of this draft. \\

\noindent\textbf{Detailed results of Experiment 2:} The detailed results of Experiment 2 (of the main paper) are provided in Tables~\ref{table:multilingual-11-1-detailed}, \ref{table:multilingual-11-2-detailed}, \ref{table:multilingual-11-3-detailed}, and \ref{table:multilingual-11-4-detailed}. The languages are from different families such as Latin (Spanish, French, Italian, Portuguese), Germanic (Danish, German, English, Dutch), and Slavic (Czech, Polish, Russian).

\begin{table*}[t]\centering
	\setlength{\tabcolsep}{5pt}
	{\small
		\centering
		\begin{tabular}{llrrrrr}
			\toprule
			
			src & trg & BilingUnsup & UMML-GW & UMML-SL & UMWE & UMH \\
			\midrule
			de & en & 70.8 & 75.7 & 74.0 & 71.8 & 75.3 \\
			de & es & 59.2 & 68.3 & 68.1 & 68.9 & 68.3 \\
			de & fr & 61.9 & 73.0 & 75.5 & 76.6 & 73.3 \\
			de & it & 60.1 & 70.0 & 72.3 & 72.2 & 68.6 \\
			de & pt & 52.6 & 59.5 & 62.9 & 62.5 & 60.6 \\
			en & de & 72.0 & 75.3 & 75.1 & 76.0 & 75.7 \\
			en & es & 80.1 & 81.7 & 81.7 & 82.9 & 82.1 \\
			en & fr & 79.8 & 82.5 & 82.5 & 83.1 & 82.1 \\
			en & it & 76.0 & 79.5 & 78.6 & 79.0 & 77.9 \\
			en & pt & 76.4 & 82.0 & 81.9 & 82.0 & 81.6 \\
			es & de & 58.9 & 68.3 & 69.2 & 68.5 & 68.4 \\
			es & en & 79.5 & 85.7 & 84.1 & 81.7 & 84.1 \\
			es & fr & 78.3 & 84.7 & 85.9 & 87.1 & 85.8 \\
			es & it & 78.7 & 82.0 & 82.8 & 84.7 & 83.5 \\
			es & pt & 82.5 & 85.5 & 86.3 & 87.8 & 87.1 \\
			fr & de & 61.1 & 69.7 & 70.8 & 70.1 & 70.0 \\
			fr & en & 78.7 & 84.4 & 83.0 & 81.0 & 83.1 \\
			fr & es & 75.6 & 81.5 & 82.9 & 84.4 & 82.7 \\
			fr & it & 75.4 & 81.6 & 82.7 & 83.3 & 83.1 \\
			fr & pt & 72.6 & 77.4 & 79.0 & 80.3 & 78.0 \\
			it & de & 59.0 & 68.5 & 69.5 & 68.6 & 67.7 \\
			it & en & 75.0 & 80.4 & 79.7 & 75.3 & 79.5 \\
			it & es & 83.6 & 86.6 & 87.2 & 88.4 & 87.4 \\
			it & fr & 81.3 & 86.6 & 87.0 & 88.3 & 87.7 \\
			it & pt & 77.1 & 79.4 & 80.9 & 82.5 & 80.9 \\
			pt & de & 56.9 & 63.3 & 65.0 & 64.2 & 63.3 \\
			pt & en & 76.5 & 82.2 & 82.1 & 77.1 & 81.5 \\
			pt & es & 87.4 & 90.5 & 91.4 & 93.1 & 91.3 \\
			pt & fr & 78.3 & 83.1 & 84.5 & 85.6 & 85.1 \\
			pt & it & 77.6 & 80.4 & 81.4 & 82.8 & 81.7 \\
			\midrule
			\multicolumn{2}{l}{average} & 72.8 & 78.3 & 79.0 & 79.0 & 78.6 \\
			\bottomrule
		\end{tabular}
	}
	\caption{
		Precision$@1$ for BLI obtained by multilingual algorithms on six European languages (German, English, Spanish, French, Italian, and Portuguese) from the MUSE dataset. 
		The `average' row reports the average performance of each method. 
	}\label{table:multilingual-european-detailed}
\end{table*}

\begin{table*}[t]\centering
	\setlength{\tabcolsep}{5pt}
	{\small
		\centering
		\begin{tabular}{llrrrrr}
			\toprule
			src & trg & BilingUnsup & UMML-GW & UMML-SL & UMWE & UMH \\
			\midrule
			cs & da & 56.3 & 60.8 & 61.0 & 62.4 & 58.8 \\
			cs & de & 66.1 & 66.6 & 67.7 & 65.5 & 65.9 \\
			cs & en & 60.8 & 64.5 & 65.9 & 61.1 & 65.1 \\
			cs & es & 66.1 & 69.6 & 69.8 & 68.7 & 68.3 \\
			cs & fr & 63.5 & 67.4 & 67.8 & 66.2 & 65.8 \\
			cs & it & 60.0 & 63.7 & 64.2 & 63.9 & 62.5 \\
			cs & nl & 62.7 & 65.8 & 66.3 & 0.0 & 63.9 \\
			cs & pl & 64.0 & 64.2 & 64.7 & 64.6 & 65.1 \\
			cs & pt & 65.0 & 68.7 & 68.9 & 68.7 & 66.8 \\
			cs & ru & 53.4 & 54.4 & 54.3 & 54.5 & 54.7 \\
			da & cs & 49.1 & 49.6 & 49.8 & 50.5 & 48.9 \\
			da & de & 69.5 & 71.0 & 70.6 & 69.6 & 70.6 \\
			da & en & 61.8 & 68.5 & 66.2 & 62.1 & 66.9 \\
			da & es & 65.2 & 68.9 & 68.8 & 67.7 & 68.0 \\
			da & fr & 61.4 & 66.0 & 65.4 & 64.5 & 64.0 \\
			da & it & 60.7 & 63.2 & 62.7 & 63.7 & 62.2 \\
			da & nl & 67.3 & 71.0 & 70.2 & 0.0 & 69.5 \\
			da & pl & 48.2 & 49.3 & 48.8 & 51.2 & 49.2 \\
			da & pt & 63.6 & 68.8 & 68.5 & 68.8 & 67.6 \\
			da & ru & 40.4 & 40.4 & 42.1 & 42.5 & 41.5 \\
			de & cs & 51.1 & 52.8 & 52.9 & 53.5 & 51.5 \\
			de & da & 63.0 & 67.2 & 67.0 & 69.3 & 66.1 \\
			de & en & 70.8 & 75.0 & 74.9 & 71.4 & 73.4 \\
			de & es & 59.2 & 68.4 & 67.7 & 67.1 & 66.7 \\
			de & fr & 61.9 & 74.7 & 74.6 & 74.6 & 72.6 \\
			de & it & 60.1 & 71.6 & 71.7 & 71.4 & 68.5 \\
			de & nl & 71.0 & 73.4 & 73.4 & 0.0 & 72.8 \\
			de & pl & 51.0 & 52.0 & 51.5 & 52.7 & 52.0 \\
			de & pt & 52.6 & 61.3 & 61.5 & 62.2 & 59.4 \\
			de & ru & 43.6 & 44.7 & 45.4 & 46.1 & 45.3 \\
			\bottomrule
		\end{tabular}
	}
	\caption{
		Precision$@1$ for BLI on eleven European languages from the MUSE dataset (part 1). 
	}\label{table:multilingual-11-1-detailed}
\end{table*}

\begin{table*}[t]\centering
	\setlength{\tabcolsep}{5pt}
	{\small
		\centering
		\begin{tabular}{llrrrrr}
			\toprule
			src & trg & BilingUnsup & UMML-GW & UMML-SL & UMWE & UMH \\
			\midrule
			en & cs & 47.1 & 54.9 & 54.8 & 55.4 & 52.2 \\
			en & da & 50.5 & 61.1 & 61.0 & 65.7 & 59.0 \\
			en & de & 72.0 & 75.1 & 75.4 & 75.2 & 75.3 \\
			en & es & 80.1 & 81.7 & 82.1 & 82.7 & 82.5 \\
			en & fr & 79.8 & 82.7 & 82.3 & 83.2 & 81.9 \\
			en & it & 76.0 & 78.9 & 78.8 & 78.3 & 77.5 \\
			en & nl & 70.0 & 76.7 & 76.1 & 0.0 & 75.3 \\
			en & pl & 53.3 & 59.2 & 59.9 & 58.5 & 55.7 \\
			en & pt & 76.4 & 81.3 & 81.7 & 81.7 & 81.3 \\
			en & ru & 44.2 & 48.8 & 50.1 & 50.7 & 46.9 \\
			es & cs & 48.9 & 50.2 & 51.0 & 53.1 & 50.6 \\
			es & da & 57.7 & 63.7 & 62.8 & 66.4 & 62.3 \\
			es & de & 58.9 & 69.1 & 69.2 & 67.4 & 67.4 \\
			es & en & 79.5 & 84.6 & 84.5 & 81.1 & 83.9 \\
			es & fr & 78.3 & 85.3 & 85.6 & 85.7 & 85.8 \\
			es & it & 78.7 & 82.5 & 82.7 & 83.5 & 83.3 \\
			es & nl & 67.0 & 70.4 & 70.5 & 0.0 & 69.3 \\
			es & pl & 51.2 & 52.7 & 52.5 & 54.8 & 52.7 \\
			es & pt & 82.5 & 86.0 & 86.1 & 86.4 & 86.5 \\
			es & ru & 46.9 & 48.1 & 48.5 & 50.4 & 47.0 \\
			fr & cs & 48.8 & 50.2 & 50.8 & 52.8 & 50.1 \\
			fr & da & 56.8 & 62.4 & 61.9 & 64.8 & 60.7 \\
			fr & de & 61.1 & 70.2 & 70.3 & 70.3 & 69.8 \\
			fr & en & 78.7 & 84.1 & 83.7 & 80.3 & 83.1 \\
			fr & es & 75.6 & 81.9 & 82.3 & 83.0 & 82.3 \\
			fr & it & 75.4 & 82.5 & 82.4 & 82.5 & 82.9 \\
			fr & nl & 69.2 & 72.3 & 72.3 & 0.0 & 71.1 \\
			fr & pl & 49.6 & 51.7 & 51.6 & 53.7 & 51.9 \\
			fr & pt & 72.6 & 79.1 & 79.2 & 78.6 & 77.9 \\
			fr & ru & 44.9 & 46.0 & 46.8 & 48.8 & 45.4 \\
			\bottomrule
		\end{tabular}
	}
	\caption{
		Precision$@1$ for BLI on eleven European languages from the MUSE dataset (part 2). 
	}\label{table:multilingual-11-2-detailed}
\end{table*}

\begin{table*}[t]\centering
	\setlength{\tabcolsep}{5pt}
	{\small
		\centering
		\begin{tabular}{llrrrrr}
			\toprule
			src & trg & BilingUnsup & UMML-GW & UMML-SL & UMWE & UMH \\
			\midrule
			it & cs & 46.5 & 49.9 & 50.9 & 52.8 & 49.6 \\
			it & da & 57.0 & 62.1 & 61.5 & 64.9 & 60.3 \\
			it & de & 59.0 & 69.3 & 69.3 & 66.6 & 66.9 \\
			it & en & 75.0 & 80.0 & 79.5 & 74.1 & 79.5 \\
			it & es & 83.6 & 87.3 & 87.2 & 88.0 & 86.9 \\
			it & fr & 81.3 & 87.1 & 87.0 & 87.1 & 87.5 \\
			it & nl & 67.3 & 71.4 & 71.4 & 0.0 & 70.1 \\
			it & pl & 52.9 & 53.5 & 53.6 & 55.9 & 53.4 \\
			it & pt & 77.1 & 80.9 & 81.1 & 80.8 & 81.0 \\
			it & ru & 44.8 & 45.8 & 45.7 & 48.6 & 45.1 \\
			nl & cs & 51.1 & 52.8 & 53.3 & 0.0 & 52.3 \\
			nl & da & 64.3 & 69.0 & 68.6 & 0.0 & 67.9 \\
			nl & de & 77.8 & 79.5 & 79.9 & 0.0 & 79.2 \\
			nl & en & 69.7 & 76.6 & 77.6 & 0.0 & 76.3 \\
			nl & es & 71.0 & 75.0 & 74.8 & 0.0 & 73.9 \\
			nl & fr & 70.6 & 74.2 & 74.1 & 0.0 & 72.8 \\
			nl & it & 67.6 & 71.3 & 71.2 & 0.0 & 69.8 \\
			nl & pl & 52.0 & 53.3 & 53.3 & 0.0 & 51.9 \\
			nl & pt & 70.6 & 74.6 & 75.0 & 0.0 & 73.2 \\
			nl & ru & 42.2 & 44.6 & 45.7 & 0.0 & 43.8 \\
			pl & cs & 61.8 & 62.5 & 63.0 & 65.0 & 63.7 \\
			pl & da & 53.2 & 57.5 & 57.6 & 60.3 & 55.2 \\
			pl & de & 63.9 & 66.3 & 66.2 & 64.8 & 64.7 \\
			pl & en & 64.2 & 68.3 & 69.8 & 64.0 & 68.3 \\
			pl & es & 66.3 & 71.1 & 71.1 & 69.4 & 69.3 \\
			pl & fr & 63.1 & 68.7 & 69.1 & 67.4 & 66.7 \\
			pl & it & 62.4 & 66.6 & 66.7 & 65.8 & 64.6 \\
			pl & nl & 62.3 & 66.3 & 66.9 & 0.0 & 64.1 \\
			pl & pt & 65.7 & 70.9 & 71.1 & 70.1 & 67.6 \\
			pl & ru & 56.8 & 57.5 & 58.3 & 59.0 & 58.4 \\
			\bottomrule
		\end{tabular}
	}
	\caption{
	Precision$@1$ for BLI on eleven European languages from the MUSE dataset (part 3). 
	}\label{table:multilingual-11-3-detailed}
\end{table*}

\begin{table*}[t]\centering
	\setlength{\tabcolsep}{5pt}
	{\small
		\centering
		\begin{tabular}{llrrrrr}
			\toprule
			src & trg & BilingUnsup & UMML-GW & UMML-SL & UMWE & UMH \\
			\midrule
			pt & cs & 46.8 & 50.1 & 50.7 & 52.0 & 49.7 \\
			pt & da & 56.3 & 62.7 & 62.3 & 64.5 & 60.7 \\
			pt & de & 56.9 & 64.3 & 64.9 & 62.5 & 62.7 \\
			pt & en & 76.5 & 81.3 & 82.4 & 76.3 & 81.1 \\
			pt & es & 87.4 & 90.9 & 91.2 & 90.6 & 91.5 \\
			pt & fr & 78.3 & 83.9 & 84.3 & 84.0 & 83.8 \\
			pt & it & 77.6 & 81.1 & 81.2 & 81.0 & 81.5 \\
			pt & nl & 65.6 & 70.5 & 70.6 & 0.0 & 69.0 \\
			pt & pl & 52.1 & 52.0 & 52.2 & 54.7 & 52.4 \\
			pt & ru & 46.3 & 45.8 & 46.8 & 49.2 & 46.0 \\
			ru & cs & 58.3 & 58.1 & 58.4 & 59.6 & 60.2 \\
			ru & da & 50.7 & 54.3 & 55.3 & 57.1 & 53.3 \\
			ru & de & 59.6 & 61.2 & 62.0 & 61.5 & 60.6 \\
			ru & en & 59.2 & 63.5 & 65.9 & 60.5 & 63.0 \\
			ru & es & 62.9 & 67.7 & 68.2 & 67.0 & 66.6 \\
			ru & fr & 62.6 & 66.1 & 66.8 & 65.9 & 65.0 \\
			ru & it & 58.2 & 62.3 & 62.8 & 62.4 & 61.0 \\
			ru & nl & 58.2 & 62.1 & 62.9 & 0.0 & 60.5 \\
			ru & pl & 61.9 & 62.8 & 63.6 & 63.6 & 63.7 \\
			ru & pt & 61.7 & 65.9 & 67.1 & 65.6 & 64.7 \\
			\midrule
			\multicolumn{2}{l}{average}  & 62.4 & 66.3 & 66.6 & 54.0 & 65.5 \\
			\bottomrule
		\end{tabular}
	}
	\caption{
		Precision$@1$ for BLI on eleven European languages from the MUSE dataset. The languages are from different families such as Latin (Spanish, French, Italian, Portuguese), Germanic (Danish, German, English, Dutch), and Slavic (Czech, Polish, Russian). This is the fourth part of the results, the first, second, and third parts being reported in Tables~\ref{table:multilingual-11-1-detailed}, \ref{table:multilingual-11-2-detailed}, and \ref{table:multilingual-11-3-detailed}, respectively. The `average' row reports the average performance of each method, the average being computed over 110 BLI results. 
	}\label{table:multilingual-11-4-detailed}
\end{table*}

\end{document}